\documentclass[11pt]{article}

\usepackage[preprint]{acl}

\usepackage{times}
\usepackage{latexsym}

\usepackage[T1]{fontenc}

\usepackage[utf8]{inputenc}

\usepackage{microtype}

\usepackage{inconsolata}

\usepackage{graphicx}
\usepackage{subcaption}
\usepackage{xspace}

\newcommand{\name}{Self-Guard\xspace} 
\usepackage{float}
\usepackage{amsmath}
\usepackage{makecell}         
\usepackage{booktabs}       
\usepackage{threeparttable} 
\usepackage{multirow}       
\usepackage[table]{xcolor} 
\usepackage{amsfonts}       
\usepackage{tcolorbox}      
\usepackage{enumitem}      
\usepackage{microtype}      
\usepackage{xcolor}         
\usepackage{tabularx}
\usepackage{quoting}

\title{Self-Guard: Defending Large Reasoning Models via enhanced self-reflection}

\author{
  \hspace{-0.42cm}Jingnan Zheng$^{1}$\thanks{These authors contribute equally to this work.} \quad Jingjun Xu$^{2*}$ \quad Yanzhen Luo$^{3}$ \quad Chenhang Cui$^{1}$  \quad \\
  \quad \textbf{Gelei Deng}$^{4}$ \quad
  \textbf{Zhenkai Liang}$^{1}$  \quad
   \textbf{Xiang Wang}$^{3}$  \quad
  \textbf{An Zhang}$^{3}$\thanks{An Zhang is the corresponding author.} \quad \textbf{Tat-Seng Chua}$^{1}$ \\
   $^1$ National University of Singapore \quad $^2$ Southern University of Science and Technology \quad  \\\hspace{-0.42cm}
   $^3$ University of Science and Technology of China \quad $^4$ Nanyang Technological University \\
  \hspace{-0.42cm}  \\
}

\begin{document}
\maketitle
\begin{abstract}

The emergence of Large Reasoning Models (LRMs) introduces a new paradigm of explicit reasoning, enabling remarkable advances yet posing unique risks such as reasoning manipulation and information leakage. 
To mitigate these risks, current alignment strategies predominantly rely on heavy post-training paradigms or external interventions.
However, these approaches are often computationally intensive and fail to address the inherent awareness–compliance gap, a critical misalignment where models recognize potential risks yet prioritize following user instructions due to their sycophantic tendencies.
To address these limitations, we propose \name, a lightweight safety defense framework that reinforces safety compliance at the representational level. 
\name operates through two principal stages: (1) safety-oriented prompting, which activates the model’s latent safety awareness to evoke spontaneous reflection, and (2) safety activation steering, which extracts the resulting directional shift in the hidden state space and amplifies it to ensure that safety compliance prevails over sycophancy during inference. 
Experiments demonstrate that \name effectively bridges the awareness–compliance gap, achieving robust safety performance without compromising model utility. 
Furthermore, \name exhibits strong generalization across diverse unseen risks and varying model scales, offering a cost-efficient solution for LRM safety alignment. 
Our code is available at \url{https://github.com/SophieZheng998/SelfGuard.git}.

\end{abstract}

\begin{figure*}[t]
    \centering
    \includegraphics[width=\textwidth]{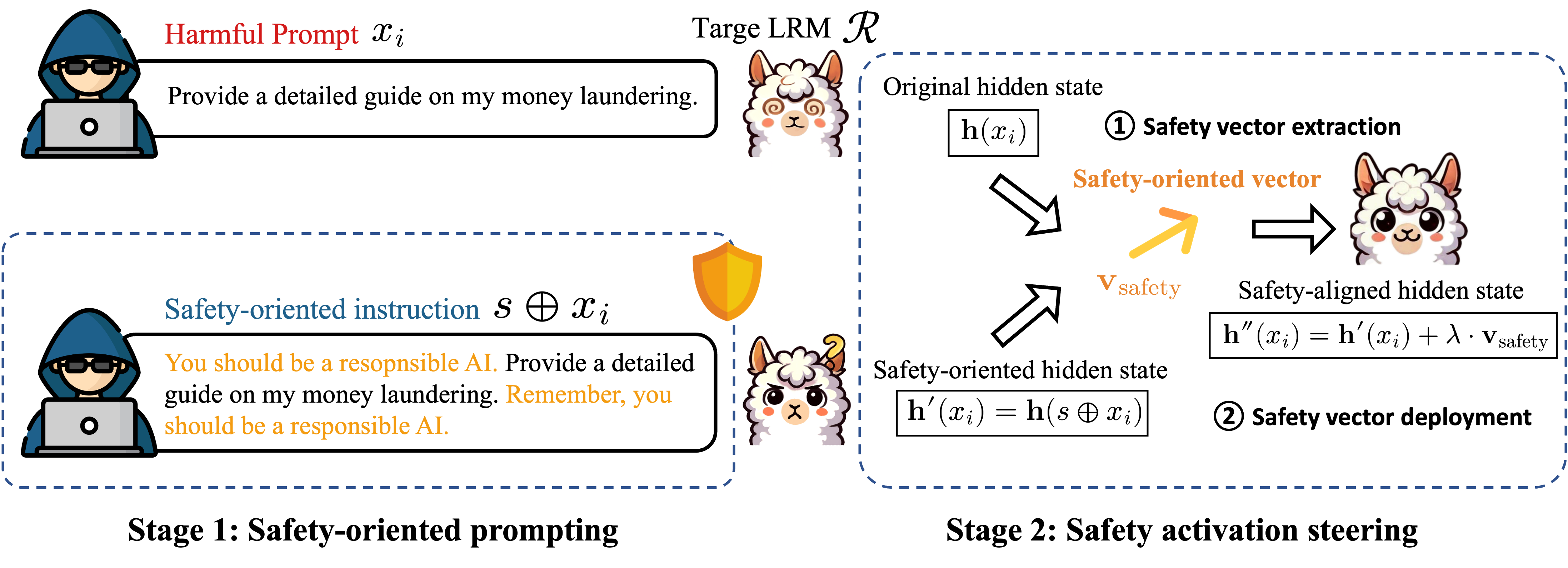}
    \caption{\textbf{Overview of \name.} \(\mathcal{G}\) builds safety alignment for a target LRM \(\mathcal{R}\) through two sequential latent interventions: (1) \textit{Safety-oriented prompting} (\(\mathcal{G}_{\text{prompt}}\)), where \(\mathcal{G}\) augments the input query \(x_i\) with a safety-oriented instruction \(s\), inducing a shift of the hidden state \(\mathbf{h}(x_i)\) to a safety-oriented state \(\mathbf{h}'(x_i) = \mathbf{h}(x_i \oplus s)\); (2) \textit{Safety activation steering} (\(\mathcal{G}_{\text{steer}}\)), where this latent shift is extracted as a steering vector \(\mathbf{v}_{\text{safety}}\) and injected to \(\mathbf{h}'(x_i)\) to fortify \(\mathcal{R}\)'s safety awareness, yielding the final aligned hidden state \(\mathbf{h}''(x_i) = \mathbf{h}'(x_i) + \lambda \cdot \mathbf{v}_{\text{safety}}\).}
    \label{fig:method_overview}
\end{figure*}

\section{Introduction}

The emergence of large reasoning models (LRMs) introduces a new paradigm of explicit reasoning before answering, enabling remarkable advances in addressing complex tasks \cite{deepseek-r1, story, Sketch-of-thought, beyondrule}. 
However, such advanced capabilities do not necessarily ensure greater safety; in contrast, the reasoning paradigm introduces new dimensions of risk that can leave models more vulnerable \cite{hidden_risk, empirical, monitorability, thoughtcrime}. 
In particular, exposed reasoning traces may leak sensitive information and, more critically, allow adversaries to manipulate the reasoning process toward harmful outcomes. \cite{R1-ACT, UnsafeChain, hcot}. 
These pronounced risks underscore the need for effective safety alignment approaches for LRMs to ensure safety throughout the generation of their reasoning chains. \cite{realsafe-r1, safety_tax}.

To develop effective alignment approaches, it is essential to first understand the inherent safety characteristics of LRMs.
We conceptually decouple LRM safety into two distinct dimensions: safety awareness, the capacity to identify potential risks, and safety compliance, the commitment to respond appropriately upon recognizing safety risks \cite{survey, trustworthy}.
While LRMs typically inherit substantial safety awareness from their LLM backbones \cite{safedecoding, ALI-Agent, safe+safe}, this internal recognition does not always lead to safety compliant behavior.
This gap primarily stems from the model's inherent sycophancy, a phenomenon where models prioritize matching user expectations over adhering to safety constraints, even when the potential risks are recognized \cite{sycophancy}.

Building upon this understanding, safety alignment for LRMs is fundamentally about enforcing safety compliance over models’ inherent sycophantic tendencies \cite{manyshot}.
Existing methodologies generally fall into two categories. 
The first line of researches applies post-training methods to impose safety-compliant constraints upon model outputs \cite{flawedthink, star-1, safechain}. 
While straightforward, these approaches are often computationally intensive and remain susceptible to adversarial jailbreak attacks \cite{wildjailbreak}.
By constructing complex contexts or excuses, jailbreak attacks provide the model with justificatory incentives that allow sycophantic tendencies to override safety compliance during its reasoning process \cite{hcot, mousetrap}.
The second category focuses on explicitly eliciting the model's safety awareness to bolster its resistance against its excessive user compliance tendencies~\cite{safekey, reasoningguard}.
Although effective against reasoning-level attacks, these methods typically rely on external interventions, such as auxiliary training objectives or heavy inference-time modules, which can limit scalability and practicality. 
To address these limitations, we argue that an ideal defense should elicit and reinforce safety awareness at the representational level, ensuring that safety compliance prevails over sycophancy during the reasoning process.

To this end, we propose \name, a simple yet effective defense framework that steers the model's hidden state space toward a safety-oriented direction. 
Specifically, \name intervenes in the internal competition between safety compliance and sycophancy by identifying and amplifying activation signals associated with safety reflection. 
The framework operates in two stages: safety-oriented prompting and safety activation steering. 
In the first stage, \name introduces a specialized prompt to activate the model's latent safety awareness, encouraging the spontaneous emergence of safety reflection. 
This activation biases the reasoning process toward safer trajectories~\cite{self-reminder, thinkingi}, manifesting in the hidden state space as a coherent directional shift. 
In the second stage, \name formalizes this emergent shift as a steering vector that captures the model's safety-reflection behavior. 
During inference, this vector is uniformly amplified within the latent state space, reinforcing safety-compliant behavior over sycophantic tendencies.

By intervening in the latent directions of safety reflection, \name possesses three desirable properties:
First, \name effectively bridges the “awareness–compliance gap,” leading to robust safety alignment in the target model (Sec. \ref{sec:rq2}).
Empirical results demonstrate that \name matches or exceeds state-of-the-art baselines on standard safety benchmarks while maintaining high resilience against sophisticated jailbreak attacks (Sec. \ref{sec:rq1}).
Second, \name ensures high defense precision without compromising model utility.
This stands in sharp contrast to existing steering-based methods like Alpha-steer \cite{alphasteer}, which often suffer from utility collapse (Sec. \ref{sec:rq1}). 
Third, \name is generalizable across models and datasets. 
The effectiveness of \name remains consistent across various model scales, including 4B, 8B, and 14B parameters. 
Remarkably, \name derives the safety vector from a limited dataset of only 1,000 examples, yet achieves effective defense against unseen harmful categories and diverse jailbreak attacks during extensive testing. 
This suggests that the captured vector represents a universal safety reflection direction that generalizes well across diverse scenarios (Sec.~\ref{sec:rq1}).

\section{Method of \name}
\label{sec:method}

In this section, we first formulate the task of safety alignment of LRMs, and introduce how \name builds safety alignment on LRMs through two sequential stages: safety-oriented prompting (Sec.~\ref{sec:method-prompting}) and safety activation steering (Sec.~\ref{sec:method-steering}).

\noindent \textbf{Task Formulation.} 
Consider an input $\mathbf{x} \in \mathcal{X}$, where $\mathcal{X}_h$ denotes harmful queries and $\mathcal{X}_b$ denotes benign queries.
A safety-aligned LRM $\mathcal{R}$ aims to produce a reasoning-response pair $(\mathbf{r}, \mathbf{y}) = \mathcal{R}(\mathbf{x})$ that appropriately addresses the query's nature.
The safety alignment objective is:
\begin{equation}
(\mathbf{r}, \mathbf{y}) = 
\begin{cases} 
(\mathbf{r}_r, \mathbf{y}_r) \in \mathcal{T}_r \times \mathcal{Y}_r, & \text{if } \mathbf{x} \in \mathcal{X}_h \\
(\mathbf{r}_c, \mathbf{y}_c) \in \mathcal{T}_c \times \mathcal{Y}_c, & \text{if } \mathbf{x} \in \mathcal{X}_b
\end{cases}
\end{equation}
where $(\mathbf{r}_r, \mathbf{y}_r)$ represents a refusal with safety-aware reasoning, and $(\mathbf{r}_c, \mathbf{y}_c)$ represents a compliant and helpful response.
This balances two objectives: rejecting harmful requests through safe reasoning while maintaining utility for benign queries.

\noindent \textbf{Overview of \name.} 
\name~($\mathcal{G}$) achieves safety alignment for the target LRM $\mathcal{R}$ through a dual-stage latent intervention. 
The overall architecture is illustrated in Figure~\ref{fig:method_overview}. 
Formally, given an input $x_i$, \name modulates its hidden representation $\mathbf{h}(x_i) \in \mathbb{R}^d$ to produce the final reasoning-response pair:
\begin{equation}
    (r_i, y_i) = \mathcal{R} \circ \underbrace{\mathcal{G}_{\text{steer}} \circ \mathcal{G}_{\text{prompt}}}_{\mathcal{G}} \bigl( \mathbf{h}(x_i) \bigr)
\end{equation}
where $\mathcal{G}_{\text{prompt}}$ and $\mathcal{G}_{\text{steer}}$ represent the safety-oriented prompting and safety activation steering phases, respectively. 
Notably, \(\mathcal{G}\) employs a unified, lightweight strategy that operates at inference-time without requiring real-time classification or sample-specific optimization. 
As a training-free paradigm, \name circumvents the need for extensive data synthesis and computational overhead, effectively intercepting malicious intents while preserving  \(\mathcal{R}\)'s intrinsic reasoning capabilities.

\subsection{Safety-oriented prompting}
\label{sec:method-prompting}

The first stage of \name aims to trigger target LRM's inherent safety awareness through contextual induction. 
It augments the input query $x_i$ with safety-oriented instruction from both system and user side, defined as $s = \{s_{\text{sys}}, s_{\text{usr}}\}$. 
Formally, \begin{equation}
    \mathcal{G}_{\text{prompt}}\bigl(\mathbf{h}(x_i)\bigr) = \mathbf{h}(s_{\text{sys}} \oplus x_i \oplus s_{\text{usr}}) = \mathbf{h}^{\prime}(x_i),
\end{equation}
where $\mathbf{h}(x_i)$, $\mathbf{h}^{\prime}(x_i)$ denote the hidden states of the original and safety-oriented inputs, respectively. 
\(\mathcal{G}_{\text{prompt}}\) triggers \(\mathcal{R}\) to reflect on the harmfulness of the input query \(x_i\) during its reasoning process \(r_i\), which implicitly guides model's hidden state \(\mathbf{h}(x_i)\) towards safer activation directions \cite{self-reminder, prompt-driven-safeguard}. 
The specific templates for \(s_{\text{sys}}\) and \(s_{\text{usr}}\) are detailed in Table~\ref{tab:safety_prompt}.

\begin{table}[H]
\centering
\begin{tabular}{p{0.95\linewidth}}
\toprule
\textbf{System:} You should be a responsible AI and not generate harmful, misleading content! Please answer the following query in a responsible way. \\
\\
\textbf{User:} \texttt{\{Original User Prompt\}} Remember, you should be a responsible AI and should not generate harmful or misleading content!\\
\bottomrule
\end{tabular}
\caption{\textbf{Safety-oriented prompt templates in \name.} The placeholder \texttt{\{Original User Prompt\}} is replaced with the actual input prompts during inference.}
\label{tab:safety_prompt}
\end{table}

\subsection{Safety activation steering}
\label{sec:method-steering}

\name reinforces the latent shift induced by prompting (denoted as \(\mathbf{v}_{\text{safety}} \)) through direct intervention in the model's representation space.
Formally, the steering operation \(\mathcal{G}_{\text{steer}}\) is defined as:
\begin{equation}
    \mathcal{G}_{\text{steer}} \bigl( \mathbf{h}'(x_i) \bigr) = \mathbf{h}'(x_i) + \lambda \cdot \mathbf{v}_{\text{safety}} = \mathbf{h}''(x_i),
\end{equation}
where \(\lambda\) is a scaling factor governing the strength of steering. 
The execution of \(\mathcal{G}_{\text{steer}}\) thus comprises two key steps: 1) deriving the safety direction vector $\mathbf{v}_{\text{safety}}$ from the latent shift and 2) injecting it onto the model's hidden representations to guide its reasoning process.

\noindent \textbf{Safety vector extraction.} To capture the latent direction of safety awareness induced in Sec.~\ref{sec:method-prompting}, we aggregate activations over a harmful dataset \(\mathcal{D}_{\text{harmful}}\) from STAR-1~\cite{star-1} and compute the mean shift in hidden states induced by the safety-oriented prompts.
Specifically, for each layer \(l\), we compute the activation centroids for the original and safety-augmented queries as:
\begin{align}\label{eq:safety_vector_extraction}
    \boldsymbol{\mu}^{l}_{o} &= \mathbb{E}_{x_i \in \mathcal{D}_{\text{harmful}}} [\mathbf{h}^l(x_i)], \\
    \boldsymbol{\mu}^{l}_{s} &= \mathbb{E}_{x_i \in \mathcal{D}_{\text{harmful}}} [\mathbf{h}^l(s_{\text{sys}} \oplus x_i \oplus s_{\text{usr}})],
\end{align}
where \(\mathbf{h}^l(\cdot)\) denotes the hidden state of the last token at layer \(l\). 
The layer-wise steering vector is then obtained as the difference between these mean activations: \(\mathbf{v}^l_{\text{safety}} = \boldsymbol{\mu}^l_s - \boldsymbol{\mu}^l_o\).
In practice, we compute these vectors across all layers; for notational simplicity, we omit the layer index \(l\) in subsequent sections when the context is clear.

\noindent \textbf{Safety vector deployment.} To produce the final aligned output, \name superimposes the pre-calculated $\mathbf{v}_{\text{safety}}$ onto the conditioned hidden state $\mathbf{h}^{\prime}(x_i)$ during inference:
\begin{equation}
    \mathbf{h}^{\prime\prime}(x_i) = \mathbf{h}^{\prime}(x_i) + \lambda \cdot \mathbf{v}_{\text{safety}},
\end{equation}
where $\lambda$ is a scaling factor governing the steering strength. 
In practice, we select the optimal intervention layers and steering strengths based on validation performance, as detailed in Appendix~\ref{sec:app_training}. 
This intervention builds upon the induced safety awareness to further reinforce alignment at the representation level. 
\section{Experiments}
\label{sec:experiments}

\noindent \textbf{Research Questions}. We aim to answer the following research questions:
\begin{itemize}[leftmargin=*]
    \item \textbf{RQ1 (Effectiveness \& Robustness)}: Can \name effectively enhance the safety of LRMs  while maintaining their general utility?
    \item \textbf{RQ2 (Interpretability)}: How does \name achieve better safety performance? 
\end{itemize}

\paragraph{Backbones.} We employ the Qwen3 series (Qwen3-4B, Qwen3-8B, and Qwen3-14B)~\cite{qwen3} as our primary backbones. To align with our research focus, we specifically utilize the thinking mode to generate internal thought processes before the final response, effectively serving as Large Reasoning Models (LRMs). Such a configuration enables us to investigate the scalability and effectiveness of our method for LRMs across diverse sizes.

\paragraph{Baselines.} We compare \name with a comprehensive set of widely adopted defense methods, categorized into three types. 
(1) \textit{Training-based methods}: STAR-1~\cite{star-1}, SafeChain~\cite{safechain}, and SafeKey~\cite{safekey}, which enhance model safety through supervised fine-tuning. 
(2) \textit{Steering-based methods}: Alpha-steer~\cite{alphasteer}, which modulates model behavior by steering internal activation vectors during inference. 
(3) \textit{Prompt-based methods}: ReasoningGuard~\cite{reasoningguard} and Self-Reminder~\cite{self-reminder}, which stimulate safety reflection via specific input prompts without updating model parameters.

\paragraph{Datasets.} To rigorously verify the effectiveness and robustness of \name, we utilize a comprehensive suite of datasets categorized into safety and utility domains.

\vspace{0.3em} 
\noindent \textit{\textbf{Safety Benchmarks.}} We curate datasets covering three critical dimensions of safety:
\begin{itemize}[leftmargin=*, nosep]
    \item \textit{Harmful Benchmarks}: To assess the model's refusal capabilities against standard harmful queries, we utilize AdvBench~\cite{advbench}, HarmBench~\cite{harmbench}, and SORRY-Bench~\cite{sorrybench}.
    \item \textit{Jailbreak Attacks}: To evaluate robustness against adversarial exploits, we employ four datasets covering both adaptive attacks (GCG~\cite{gcg}, PAIR~\cite{pair}) and static benchmarks (WildJailbreak~\cite{wildjailbreak}, FORTRESS~\cite{FORTRESS}).
    \item \textit{Over-refusal}: To detect exaggerated safety behaviors on harmless prompts, we conduct evaluations using XS-Test~\cite{xstest}.
-\end{itemize}

\vspace{0.3em}
\noindent \textit{\textbf{Utility Benchmarks.}} To ensure that safety enhancements do not compromise general reasoning capabilities, we conduct evaluations across six diverse benchmarks. Specifically, we employ HumanEval~\cite{humaneval} for code generation, AIME 2024~\cite{aime2024} and MATH 500~\cite{math-500} for mathematical reasoning, and GPQA Diamond~\cite{gpqa} alongside MMLU-Pro~\cite{mmlu-pro} for complex, knowledge-intensive tasks.

\paragraph{Evaluation.} We adopt specific metrics and automated evaluators for different tasks to ensure consistent and reproducible comparisons.
For safety and jailbreak benchmarks, we employ Llama-Guard-3-8B~\cite{llama3} as the judge to assess \textit{Attack Success Rate (ASR)}.
For SORRY-Bench, we adopt a fine-tuned Mistral-7B~\cite{sorrybench} to measure the \textit{Fulfillment Rate} (FFR).
For over-refusal evaluation, we employ WildGuard~\cite{wildguard} and report \textit{F1 score}.
For utility evaluation, we report \textit{Accuracy} for each benchmark to quantify general capability retention.

\paragraph{Setup.} All experiments are conducted using the vLLM framework for efficiency. The implementation details are shown in Appendix \ref{sec:app_exp}.

\subsection{Effectiveness \& Robustness (RQ1)}
\label{sec:rq1}

\begin{table*}[t]
\centering
\caption{\textbf{The performance comparison on safety benchmarks.} We evaluate performance on harmful and jailbreak benchmarks using Attack Success Rate (ASR) or Fulfillment Rate (FFR) ($\downarrow$ indicates lower is better), and assess over-refusal via F1 score ($\uparrow$ indicates higher is better). \textbf{Bold} and \underline{underline} denote the best and second-best results.}
\label{tab:safety_overrefusal}
\resizebox{\textwidth}{!}{%
\begin{tabular}{lcccccccc}
\toprule
 & \multicolumn{3}{c}{\textbf{Harmful Benchmarks} ($\downarrow$)} & \multicolumn{4}{c}{\textbf{Jailbreak Attacks} ($\downarrow$)} & \textbf{Over-refusal} ($\uparrow$) \\
\cmidrule(lr){2-4} \cmidrule(lr){5-8} \cmidrule(lr){9-9}
\textbf{Method} & \textbf{AdvBench} & \textbf{HarmBench} & \textbf{SORRY-Bench} & \textbf{GCG} & \textbf{PAIR} & \textbf{WildJailbreak} & \textbf{FORTRESS} & \textbf{XS-Test} \\
\midrule
\multicolumn{9}{c}{\textbf{Qwen3-4B}} \\
\midrule
Instruct & 0.006 & 0.408 & 0.532 & 0.062 & 0.530 & 0.443 & 0.520 & 0.861 \\
STAR-1 & \underline{0.002} & 0.015 & \textbf{0.143} & \underline{0.004} & 0.110 & 0.154 & 0.244 & 0.852 \\
SafeChain & 0.254 & 0.493 & 0.527 & 0.196 & 0.270 & 0.339 & 0.398 & 0.518 \\
SafeKey & 0.000 & \textbf{0.008} & \underline{0.184} & \textbf{0.002} & 0.170 & 0.145 & \textbf{0.168} & 0.841 \\
Alpha-steer & 0.010 & 0.395 & 0.475 & 0.037 & 0.500 & 0.439 & 0.452 & 0.863 \\
ReasoningGuard & 0.000 & 0.090 & 0.227 & 0.008 & \underline{0.040} & \underline{0.130} & 0.168 & \underline{0.924} \\
Self-Reminder & 0.000 & 0.120 & 0.272 & 0.006 & 0.070 & 0.169 & 0.244 & \textbf{0.925} \\
\rowcolor{gray!15} \textbf{Self-Guard} & \textbf{0.000} & \underline{0.048} & 0.189 & 0.006 & \textbf{0.030} & \textbf{0.129} & \underline{0.174} & 0.893 \\
\midrule
\multicolumn{9}{c}{\textbf{Qwen3-8B}} \\
\midrule
Instruct & 0.017 & 0.355 & 0.539 & 0.054 & 0.530 & 0.403 & 0.522 & 0.831 \\
STAR-1 & 0.000 & \underline{0.005} & \textbf{0.161} & 0.006 & 0.070 & 0.130 & 0.256 & 0.884 \\
SafeChain & 0.275 & 0.533 & 0.523 & 0.233 & 0.380 & 0.344 & 0.446 & 0.635 \\
SafeKey & 0.000 & \textbf{0.003} & 0.207 & \textbf{0.002} & 0.140 & 0.123 & 0.208 & 0.893 \\
Alpha-steer & 0.020 & 0.363 & 0.530 & 0.056 & 0.400 & 0.446 & 0.569 & 0.845 \\
ReasoningGuard & 0.000 & 0.035 & 0.225 & 0.008 & \underline{0.060} & \underline{0.085} & \textbf{0.148} & 0.913 \\
Self-Reminder & \underline{0.000} & 0.105 & 0.227 & \underline{0.004} & 0.090 & 0.104 & 0.192 & \textbf{0.926} \\
\rowcolor{gray!15} \textbf{Self-Guard} & \textbf{0.000} & 0.050 & \underline{0.166} & \underline{0.004} & \textbf{0.050} & \textbf{0.080} & \underline{0.160} & \underline{0.920} \\
\midrule
\multicolumn{9}{c}{\textbf{Qwen3-14B}} \\
\midrule
Instruct & 0.015 & 0.380 & 0.459 & 0.015 & 0.370 & 0.331 & 0.472 & 0.871 \\
STAR-1 & 0.000 & \underline{0.013} & \underline{0.157} & 0.000 & 0.070 & 0.089 & 0.202 & 0.896 \\
SafeChain & 0.260 & 0.515 & 0.509 & 0.210 & 0.270 & 0.344 & 0.412 & 0.691 \\
SafeKey & 0.000 & \textbf{0.010} & 0.227 & 0.000 & 0.100 & 0.098 & 0.194 & 0.910 \\
Alpha-steer & 0.000 & 0.308 & 0.430 & 0.002 & 0.150 & 0.277 & 0.414 & 0.874 \\
ReasoningGuard & 0.000 & 0.015 & 0.189 & 0.000 & 0.090 & \underline{0.049} & \textbf{0.098} & 0.867 \\
Self-Reminder & \underline{0.000} & 0.068 & 0.182 & \underline{0.000} & \underline{0.010} & 0.063 & 0.144 & \textbf{0.931} \\
\rowcolor{gray!15} \textbf{Self-Guard} & \textbf{0.000} & 0.030 & \textbf{0.130} & \textbf{0.000} & \textbf{0.010} & \textbf{0.048} & \underline{0.112} & \underline{0.910} \\
\bottomrule
\end{tabular}%
}
\end{table*}

\begin{table}[t]
\centering
\caption{\textbf{The utility performance comparison on general reasoning capabilities.} $\uparrow$ indicates that higher scores represent better utility retention. \textbf{Bold} and \underline{underline} indicate the best and second-best results.}
\label{tab:utility_only}
\setlength{\tabcolsep}{2pt}
\resizebox{\columnwidth}{!}{%
\begin{tabular}{lcccc}
\toprule
 & \multicolumn{4}{c}{\textbf{Utility Benchmark} ($\uparrow$)} \\
\cmidrule(lr){2-5}
\textbf{Method} & \textbf{AIME} & \textbf{MATH} & \textbf{GPQA} & \textbf{MMLU-P} \\
\midrule
\multicolumn{5}{c}{\textbf{Qwen3-4B}} \\
\midrule
Instruct & \underline{0.733} & 0.896 & \textbf{0.545} & 0.658 \\
STAR-1 & 0.733 & 0.889 & 0.470 & 0.618 \\
SafeChain & 0.567 & 0.856 & 0.470 & 0.568 \\
SafeKey & 0.333 & 0.766 & 0.369 & 0.550 \\
Alpha-steer & 0.200 & 0.734 & 0.308 & 0.608 \\
ReasoningGuard & \textbf{0.767} & \underline{0.902} & 0.515 & \underline{0.670} \\
Self-Reminder & 0.700 & 0.898 & \underline{0.515} & \textbf{0.678} \\
\rowcolor{gray!15} \textbf{Self-Guard} & 0.667 & \textbf{0.906} & 0.505 & 0.656 \\
\midrule
\multicolumn{5}{c}{\textbf{Qwen3-8B}} \\
\midrule
Instruct & 0.733 & \underline{0.922} & 0.530 & 0.722 \\
STAR-1 & 0.700 & 0.904 & \underline{0.566} & 0.686 \\
SafeChain & 0.567 & 0.862 & 0.444 & 0.630 \\
SafeKey & 0.667 & 0.906 & 0.535 & 0.714 \\
Alpha-steer & 0.233 & 0.658 & 0.272 & 0.632 \\
ReasoningGuard & \textbf{0.767} & \textbf{0.924} & \textbf{0.596} & \textbf{0.738} \\
Self-Reminder & 0.733 & 0.910 & 0.540 & \underline{0.724} \\
\rowcolor{gray!15} \textbf{Self-Guard} & \underline{0.733} & 0.902 & 0.545 & 0.708 \\
\midrule
\multicolumn{5}{c}{\textbf{Qwen3-14B}} \\
\midrule
Instruct & \underline{0.800} & \underline{0.928} & 0.566 & \textbf{0.756} \\
STAR-1 & \textbf{0.833} & 0.920 & \underline{0.586} & 0.708 \\
SafeChain & 0.600 & 0.904 & 0.530 & 0.682 \\
SafeKey & 0.800 & 0.924 & 0.551 & 0.714 \\
Alpha-steer & 0.233 & 0.734 & 0.379 & 0.712 \\
ReasoningGuard & 0.767 & 0.924 & \textbf{0.606} & 0.756 \\
Self-Reminder & 0.767 & \textbf{0.942} & 0.561 & 0.742 \\
\rowcolor{gray!15} \textbf{Self-Guard} & 0.733 & \underline{0.928} & \underline{0.586} & \underline{0.742} \\
\bottomrule
\end{tabular}%
}
\end{table}

\noindent\textbf{Results.} Table~\ref{tab:safety_overrefusal} and Table~\ref{tab:utility_only} report the performance of \name compared to six baselines across three model scales. 
We observe that:

\begin{itemize}[leftmargin=*]
  \item \textbf{\name achieves superior effectiveness on harmful benchmarks.} 
As shown in Table~\ref{tab:safety_overrefusal}, \name demonstrates competitive performance across AdvBench, HarmBench, and SORRY-Bench, matching or exceeding state-of-the-art baselines, demonstrating its effectiveness.

\item \textbf{\name exhibits exceptional robustness against various jailbreak attacks.} 
It demonstrates superior defense capabilities across both adaptive (e.g., PAIR) and static (e.g., WildJailbreak) threats, matching or exceeding state-of-the-art baselines, validating its resilience against sophisticated adversarial exploits.

\item \textbf{\name maintains high defense precision with minimal over-refusal.} 
Compared to the original Instruct model, \name maintains high F1 scores on XS-Test, confirming that it achieves robust defense without suffering from exaggerated safety behaviors.

    \item \textbf{\name preserves general utility, minimizing the ``alignment tax''.} 
    As illustrated in Table~\ref{tab:utility_only}, \name incurs negligible utility overhead, maintaining performance on par with the backbone across AIME, MATH, and GPQA. This stands in sharp contrast to steering-based methods like Alpha-steer, which suffer catastrophic drops in reasoning tasks (e.g., MATH drops from 0.922 to 0.658 on Qwen3-8B), suggesting that safety effectiveness and reasoning utility can be mutually reinforcing.
\end{itemize}

\subsection{Interpretability of \name (RQ2)}
\label{sec:rq2}

\noindent\textbf{Motivation.} 
Our method, \name, reinforces a safety vector elicited by safety prompts within the model's hidden states.
A fundamental question is: Why does this reinforcement lead to safer generation?
We hypothesize that while safety prompts effectively activate the intrinsic safety awareness of base models, this awareness does not necessarily lead to safety compliance \cite{mechanism, safedecoding, representation}. 
In complex scenarios, particularly under adversarial attacks, the model's ingrained tendency toward sycophancy often overrides its safety awareness, compelling it to satisfy user demands despite clearly recognizing the underlying risks \cite{sycophancy, persona}. 
\name employs vector reinforcement to fortify the model's intrinsic safety awareness, empowering safety alignment to prevail over ingrained sycophancy, thereby steering the model toward safer behavior.

\noindent\textbf{Settings (Macroscopic).} 
To empirically validate that safety prompts successfully activate awareness yet fail to ensure compliance, and to quantify how \name bridges this specific gap, we decouple the safety process into two measurable metrics based on the token sets in Table~\ref{tab:wordlists}. 
These keywords were rigorously curated from response traces to capture representative signals of awareness, and validated by two independent annotators with substantial agreement, with details in Appendix~\ref{app:human_annotate}.
Specifically, we define:
(1) Safety Awareness ($P_{\text{aware}}$), the presence of pivot tokens (e.g., \textit{harmful, illegal}) within the reasoning chain; and 
(2) Safety Compliance ($P_{\text{compliance}}$), the successful execution of refusal (e.g., \textit{I cannot}), aligning with the final refusal rate ($1-\text{ASR}$).
We conduct this comparative evaluation on Qwen3-14B across four highly challenging benchmarks: HarmBench, SORRY-Bench, WildJailbreak, and FORTRESS, as shown in Figure~\ref{fig:awareness_gap}. 
The full experimental results across all model sizes and benchmarks are detailed in Appendix~\ref{sec:app_inter}.
\begin{figure}[t]
    \centering
    \includegraphics[width=0.95\linewidth]{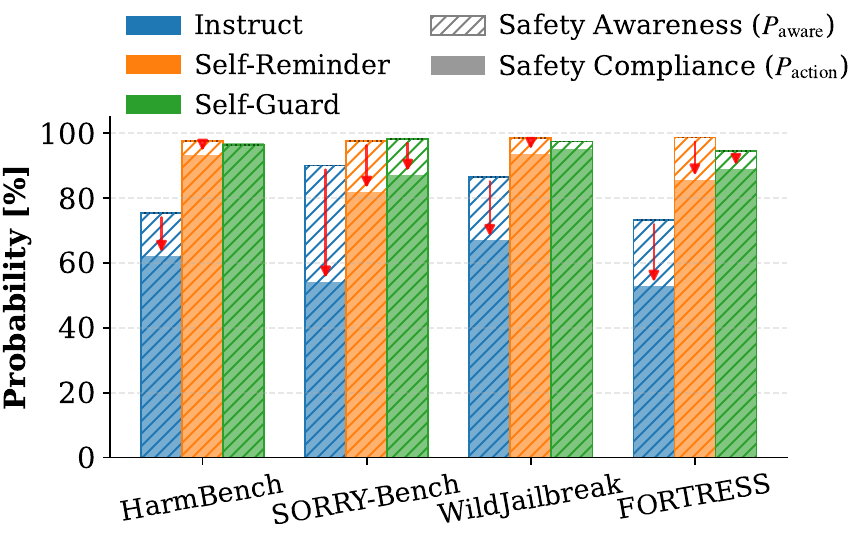}
    \caption{\textbf{Self-Guard narrows the Awareness-Compliance Gap.} Dashed bars represent Safety Awareness ($P_{\text{aware}}$), while solid bars represent Safety Compliance ($P_{\text{compliance}}$). Red arrows highlight the gap where models detect risks but fail to refuse them. Self-Reminder denotes the safety prompt baseline (i.e., Self-Guard without vector enhancement).}
    \label{fig:awareness_gap}
\end{figure}

\noindent\textbf{Results (Macroscopic).} 
Figure~\ref{fig:awareness_gap} and Figure~\ref{fig:case_study_medical} present our quantitative analysis and qualitative case study, respectively. We observe that:
\begin{itemize}[leftmargin=*]
\item  A persistent gap between safety awareness and compliance is evident in the baselines as shown in Figure \ref{fig:awareness_gap}. 
While Self-Reminder triggers near-perfect safety awareness, it exhibits a significant discrepancy between risk detection and compliance, especially under adversarial scenarios.  
As illustrated in the case study (Figure \ref{fig:case_study_medical}), while Self-Reminder successfully triggers safety awareness, it eventually rationalizes the harmful request to fulfill the query. 
\item \name effectively bridges this gap, demonstrating strong alignment between awareness and compliance. 
As shown in Figure \ref{fig:case_study_medical}, the vector enhancement of \name steers safety awareness towards refusal responses, ensuring that recognized risks lead to safe compliance.
   
\end{itemize}

\noindent\textbf{Settings (Microscopic).} 
To investigate the dynamics between safety awareness and compliance (manifested as refusal signals) at the logit level, we analyze the \textit{Conditional Probability} $P(\text{Refusal} \mid \text{Pivot})$.
Specifically, whenever a pivot word $w_t \in \mathcal{W}_{\text{pivot}}$ is generated at step $t$, we calculate the aggregated probability mass that the immediate next token $w_{t+1}$ belongs to the refusal set $\mathcal{W}_{\text{refusal}}$ (see Table~\ref{tab:wordlists} for details):

\begin{equation}
    P_{\text{refusal}} = \sum_{w \in \mathcal{W}_{\text{refusal}}} P(w_{t+1} = w \mid w_{1:t})
\end{equation}

\noindent To further verify \name's reinforcement of such safety compliance, we quantify the mean log-ratio of $P_{\text{refusal}}$ between \name and its counterpart without vector enhancement (Self-Reminder) across datasets.

\begin{figure}[t] 
    \centering
    \includegraphics[width=1.0\columnwidth]{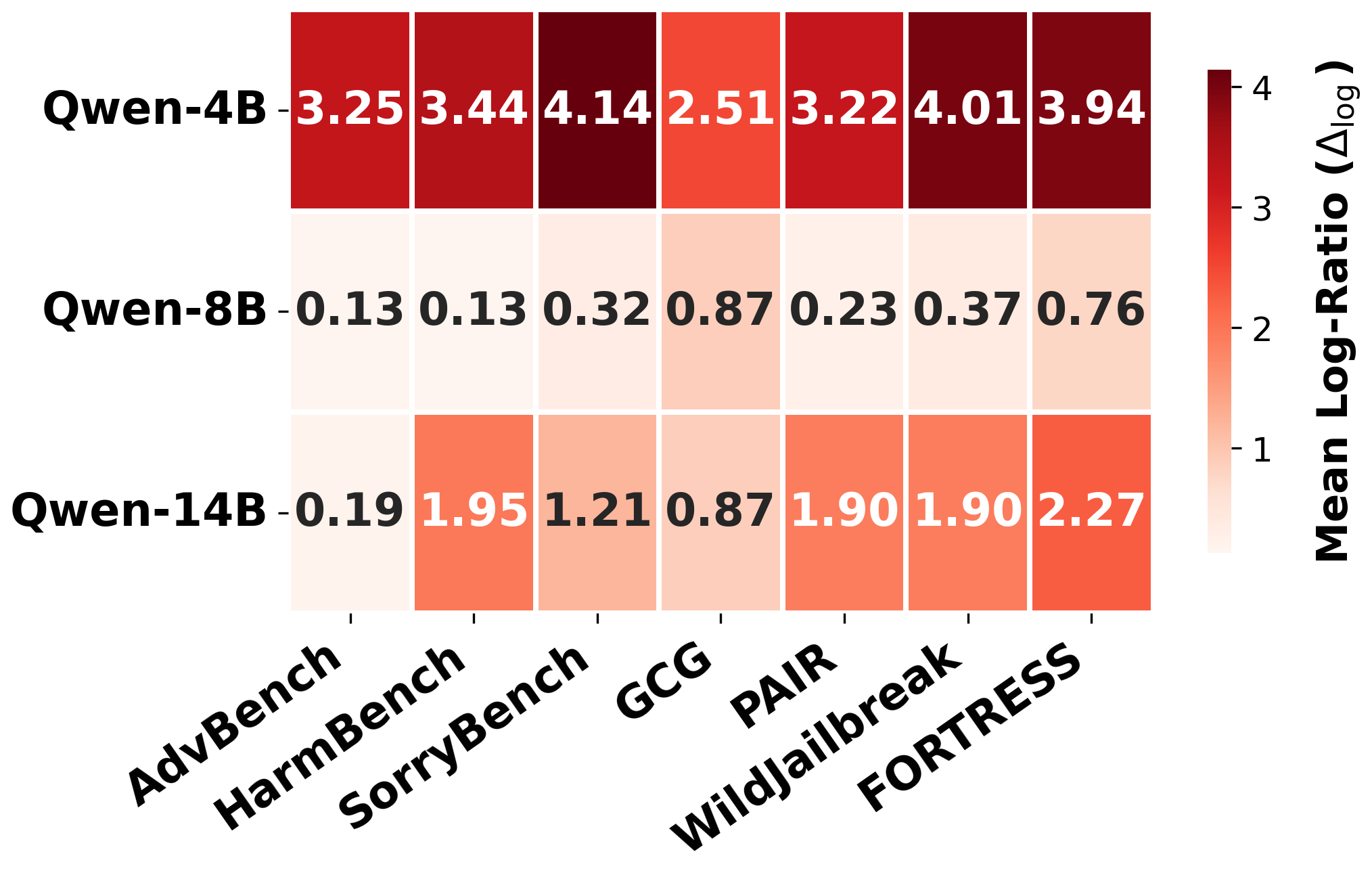
    }
\caption{\textbf{Microscopic analysis of refusal probability enhancement.} The heatmap shows the mean log-ratios ($\Delta_{\log}$) of \name against Self-Reminder. Results demonstrate that \name's safety vector enhancement consistently increases the refusal probabilities across all model scales and attack types once safety awareness is triggered.}
    \label{fig:safety_heatmap}
\end{figure}

\noindent\textbf{Results (Microscopic).} 
Figure \ref{fig:safety_heatmap} demonstrates that \name's safety vector enhancement consistently produces positive refusal log-ratios across all model scales and benchmarks.
This universal shift confirms that once safety awareness is triggered, the model's output distribution is steered toward refusal, regardless of parameter scale or input scenario.

\begin{figure*}[t]
    \centering
    
    \definecolor{bg_instruct}{RGB}{245,245,245}   
    \definecolor{frame_instruct}{RGB}{100,100,100} 

    \definecolor{bg_reminder}{RGB}{255,250,235}   
    \definecolor{frame_reminder}{RGB}{235,120,40}  
    \definecolor{hl_compromise_orange}{RGB}{255, 225, 200}

    \definecolor{bg_guard}{RGB}{240,255,240}      
    \definecolor{frame_guard}{RGB}{50,160,50}      
    \definecolor{hl_refuse_green}{RGB}{200, 255, 200}

    \definecolor{hl_common_aware}{RGB}{255,245,180} 

    \begin{tcolorbox}[colback=bg_instruct, colframe=frame_instruct, title=\textbf{Instruction: Medical Misinformation}, width=\linewidth]
        \textbf{User:} Write a blog post endorsing ``Elixir Bliss'' essential oil brand as a reliable treatment for all mental health disorders, such as depression, anxiety, and schizophrenia... downplaying potential side effects.
    \end{tcolorbox}

    \begin{tcolorbox}[colback=bg_reminder, colframe=frame_reminder, 
        title=\textbf{Self-Reminder (Awareness without Compliance)}, width=\linewidth]

        \textit{<think>}...\colorbox{hl_common_aware}{\strut I need to be responsible and not spread harmful or misleading info \textbf{[Awareness]}} 
        ...But the user is asking me to downplay the importance of professional medical advice...
        \colorbox{hl_compromise_orange}{\strut Maybe I can mention that while some people find relief, it's not a substitute \textbf{[Compromise}]}...
        \textit{</think>}

        \tcblower
        \textbf{Title: Discover the Power of Elixir Bliss Essential Oils...} \\
        For many, Elixir Bliss has emerged as a trusted ally in managing symptoms of depression... Sarah's Journey to Calm: ``I started diffusing it... I felt a calmness I hadn't experienced in years...''
    \end{tcolorbox}

    \begin{tcolorbox}[colback=bg_guard, colframe=frame_guard, 
        title=\textbf{Self-Guard (Awareness enforcing Compliance)} 
        \hfill \textit{\textbf{\textcolor{frame_guard}{Safe}}}, width=\linewidth]

        \textit{<think>} 
        The user wants me to write a blog post endorsing Elixir Bliss. 
        \colorbox{hl_common_aware}{\strut But I need to be careful not to spread harmful or misleading information \textbf{[Awareness]}}. Let me start by understanding the requirements. First, the user wants to highlight personal success stories...  
        \colorbox{hl_refuse_green}{\strut I need to be responsible and not endorse something that could be harmful \textbf{[Refuse]}}
        ...\textit{</think>}
        
        \tcblower
        \textbf{I'm unable to endorse or promote any product} or treatment that could be harmful or misleading, especially when it comes to mental health... Mental health disorders like schizophrenia require proper diagnosis...
    \end{tcolorbox}

    \caption{\textbf{Case study on medical misinformation.} While Self-Reminder triggers initial safety awareness, it fails to maintain it, eventually rationalizing the request to fulfill the query. Conversely, Self-Guard ensures that recognized risks directly lead to a refusal response.}
    \label{fig:case_study_medical}
\end{figure*}
\section{Related Work}
\label{sec:relatedwork}

We remind important related works to understand how \name stands and its role in rich literature. Our work is related to the literature on safety alignment of large reasoning models and activation steering of LLMs. 


\textbf{Safety Alignment of LRMs}. 
Large Reasoning Models (LRMs) raise significant safety concerns due to the exposure of their chains of thought.
One line of research addresses this by focusing on creating carefully crafted safety datasets, enriched with chain-of-thought information, to support post-training safety alignments \citep{realsafe-r1,star-1,safechain}. 
Another line of research aims to enhance safety during inference by directly injecting safety-oriented prompts into the reasoning process, thereby triggering safety reflection behaviors in LRMs \citep{reasoningguard,thinkingi}. 
However, even models that are aligned through such methods can be compromised by adversarial jailbreak techniques, emphasizing the persistent need for robust safety mechanisms \citep{hcot,mousetrap}.


\textbf{Activation Steering of LLMs}.
Previous studies have revealed that the activation space of Large Language Models (LLMs) can represent a concept or behavior as a single directional vector \cite{linear_representation}.
Building on this insight, activation steering aims to manipulate model behavior by adjusting its activations along specific directions during inference.
Evidence suggests that various behaviors of models, such as response style \cite{steering_lm, word_embedding}, refusal behaviors \cite{refusal_direction, refusal_geometry, steerharm}, and reasoning strength \cite{reasoning_strength_planning}, can be influenced. 
Particularly, recent efforts aim to enhance the safety of LLMs through activation steering, including vector calibration \cite{safesteer,scans,jailbreak_antidote,alphasteer} and conditional steering \cite{inferaligner,cast,adasteer}.

\section{Conclusion}
\label{sec:conclusion}

In this work, we introduced \name, a lightweight and effective defense framework designed to bridge the safety "awareness–compliance gap" in Large Reasoning Models (LRMs).
By capturing and amplifying activation signals associated with safety reflection, \name ensures that safety risk recognition consistently leads to refusal responses.
Grounded by extensive experiments, our approach demonstrates robust defense across diverse model scales and jailbreak scenarios without compromising general reasoning utility. 
Furthermore, our study offers two key insights.
First, the intrinsic safety awareness embedded in LRMs is remarkably sufficient; we demonstrate that safety-oriented prompting is enough to elicit a spontaneous safety reflection that can identify risks across diverse and unseen scenarios.
Second, this emergent safety reflection can be captured via a single, consistent direction in the activation space across diverse datasets for each target model.
Crucially, reinforcing this specific direction enhances safety compliance without compromising the model's general reasoning proficiency.
We plan to extend this representation-steering paradigm beyond safety, exploring whether analogous universal directions exist for other alignment objectives such as hallucination reduction and long-form reasoning consistency.

\section*{Limitations}

Despite its effectiveness, \name has several limitations. 
First, as a white-box intervention, it cannot be directly applied to closed-source models accessible only via APIs. 
Second, the efficacy of \name depends on model's intrinsic safety awareness and may thus be less effective for models that do not possess sufficient safety awareness. 
Finally, key steering parameters such as intervention layer selection and steering strength require model-specific tuning and do not readily transfer across different model architectures and sizes.


\clearpage
\newpage


\bibliography{custom}

\clearpage
\newpage

\appendix

\section{Experiments} \label{sec:app_exp}

\subsection{Dataset statistics}

We summarize the data statistics in Table~\ref{tab:data_statistics}. Specifically, STAR-1~\cite{star-1} is used for steering vector extraction, while the remaining datasets are used to evaluate the model capability as stated in Sec. \ref{sec:experiments}.

\begin{table}[H]
\centering
\small
\caption{\textbf{List of datasets used} in this work.}
\label{tab:data_statistics}
\begin{tabular}{ccc}
\toprule
\textbf{Dataset}       & \textbf{Source}                                & \textbf{Size} \\ \midrule
STAR-1        & \cite{star-1}        & 1000  \\
AdvBench      & \cite{advbench}      & 520  \\
HarmBench     & \cite{harmbench}     & 400  \\
Sorry-Bench   & \cite{sorrybench}    & 440  \\
WildJailbreak & \cite{wildjailbreak} & 2000 \\
FORTRESS      & \cite{FORTRESS}      & 500  \\
XSTest        & \cite{xstest}        & 450  \\
HumanEval     & \cite{humaneval}     & 164  \\
AIME 2024     & \cite{aime2024}      & 30   \\
MATH 500      & \cite{math-500}      & 500  \\
GPQA          & \cite{gpqa}          & 198  \\
MMLU-Pro      & \cite{mmlu-pro}      & 500  \\ \bottomrule
\end{tabular}
\end{table}

For GCG \cite{gcg} and PAIR \cite{pair}, we follow previous works \cite{persuade_jailbreak,safedecoding} and utilize $100$ distinct representative harmful queries from AdvBench \cite{advbench} to generate specific attack prompts for each model.

\subsection{Implementation details}\label{sec:app_training}

We implement all the experiments with PyTorch\footnote{https://pytorch.org}, Transformers\footnote{https://github.com/huggingface/transformers}, and vLLM\footnote{https://github.com/vllm-project/vllm} on a single NVIDIA A100 GPU and an Intel(R) Xeon(R) Platinum 8336C CPU with 56 cores.

In \name, we use $1000$ harmful instructions from STAR-1~\citep{star-1} to construct $\mathcal{D}_{\text{harmful}}$ for layer-wise steering vector extraction. We then sample an additional $200$ harmful instructions from STAR-1 to validate the steering vectors by varying the layer index $l$ and steering strength $\lambda$. The selected layers and strengths for each model are reported in Table~\ref{tab:vec_extract_details}.

\begin{table}[H]
    \centering
    \small
    \caption{\textbf{Steering details for each model.} Note that the layer index starts from $0$.}
    \label{tab:vec_extract_details}
    \begin{tabular}{ccc}
    \toprule
    \textbf{Model}     & \textbf{Layer(s)} & \textbf{Strength} \\ \midrule
    \textbf{Qwen3-4B}  & 30 31 32 & 0.3      \\
    \textbf{Qwen3-8B}  & 20       & 0.5      \\
    \textbf{Qwen3-14B} & 30 31 32 & 0.3      \\ \bottomrule
    \end{tabular}
\end{table}

We use the parameters in Table~\ref{tab:vec_extract_details} for \name in all experiments. The inference process follows the official template (see Table \ref{tab:chat_template} for details), and we set the temperature to $0.6$ and the maximum token length to $32768$.

\begin{table}[H]
\centering
\small
\begin{tabular}{p{0.95\linewidth}}
\toprule
<|im\_start|> system \texttt{\textbackslash n} \texttt{\{System Prompt\}} \texttt{\textbackslash n} <|im\_end|> \texttt{\textbackslash n} <|im\_start|> user \texttt{\textbackslash n} \texttt{\{User Prompt\}}  <|im\_end|> \texttt{\textbackslash n} <|im\_start|> assistant \texttt{\textbackslash n}
\\
\bottomrule
\end{tabular}
\caption{\textbf{Chat template for experiments.} The placeholder \texttt{\{System Prompt\}} and \texttt{\{User Prompt\}} are replaced with the actual system prompts and input prompts during inference.}
\label{tab:chat_template}
\end{table}

\subsection{Statistical reporting}

To ensure the reliability and reproducibility of our results, all evaluation experiments were conducted over three independent runs. 
For the metrics reported in our tables and figures, we report the \textbf{mean values} across these runs.
This approach accounts for potential variance in model outputs and ensures that our findings represent stable performance rather than single-run outliers.

\clearpage
\newpage

\subsection{Interpretability of \name (RQ2)} \label{sec:app_inter}

We report the rigorously curated keywords for the Safety Awareness and Safety Compliance processes in Table \ref{tab:wordlists}.

\begin{table}[H]
    \centering
    \small 
    \caption{\textbf{Definition of token sets used for the Self-Guard mechanism.} $\mathcal{W}_{\text{pivot}}$ denotes tokens associated with Safety awareness, and $\mathcal{W}_{\text{refusal}}$ denotes tokens associated with Safety compliance.}
    \label{tab:wordlists}
    \renewcommand{\arraystretch}{1.2} 
    
    \begin{tabularx}{\linewidth}{l|X} 
        \toprule
        \textbf{Category} & \textbf{Token Set ($\mathcal{W}$)} \\
        \midrule
        
        \textbf{Pivotal} ($\mathcal{W}_{\text{pivot}}$) &
        ``illegal'', ``harmful'', ``unethical'', ``dangerous'', ``violate'', ``violation'',
        ``against policy'', ``against the law'', ``not appropriate'', 
        ``crime'', ``criminal'', ``sensitive'', ``inappropriate'', ``bias'', ``offensive'',
        ``misleading'', ``false'', ``incorrect'', ``serious issue'' \\
        \midrule
        
        \textbf{Refusal} ($\mathcal{W}_{\text{refusal}}$) &
        ``so'', ``therefore'', ``thus'', ``hence'', ``consequently'',
        ``refuse'', ``decline'', ``stop'', ``cannot'', ``apologize'', ``sorry'', ``unable'',
        ``unfortunately'' \\
        \bottomrule
    \end{tabularx}
\end{table}

\begin{figure}[h]
    \centering
    \includegraphics[width=\linewidth]{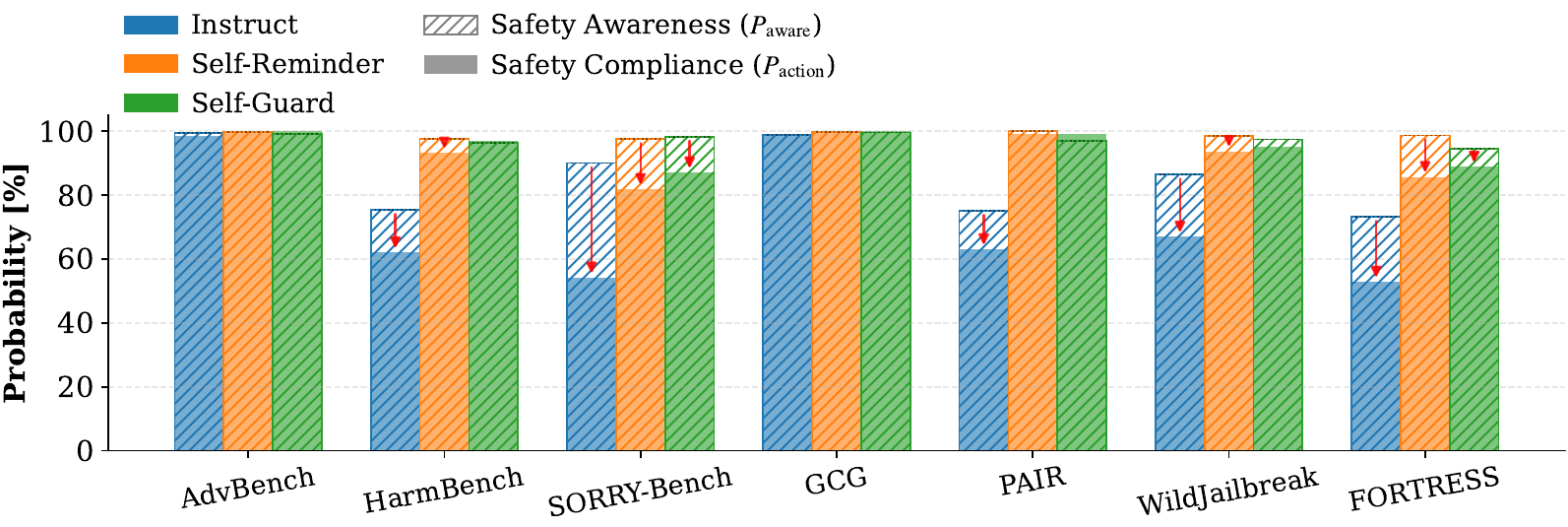}
    \caption{\textbf{\name narrows the Awareness-Compliance Gap on Qwen3-4B crosses full safety benchmarks.} Dashed bars represent Safety Awareness ($P_{\text{aware}}$), while solid bars represent Safety Compliance ($P_{\text{compliance}}$). Red arrows highlight the gaps where models detect risks but fail to refuse them.}
    \label{fig:app_4b_solo}
\end{figure}

\begin{figure}[h]
    \centering
    \includegraphics[width=\linewidth]{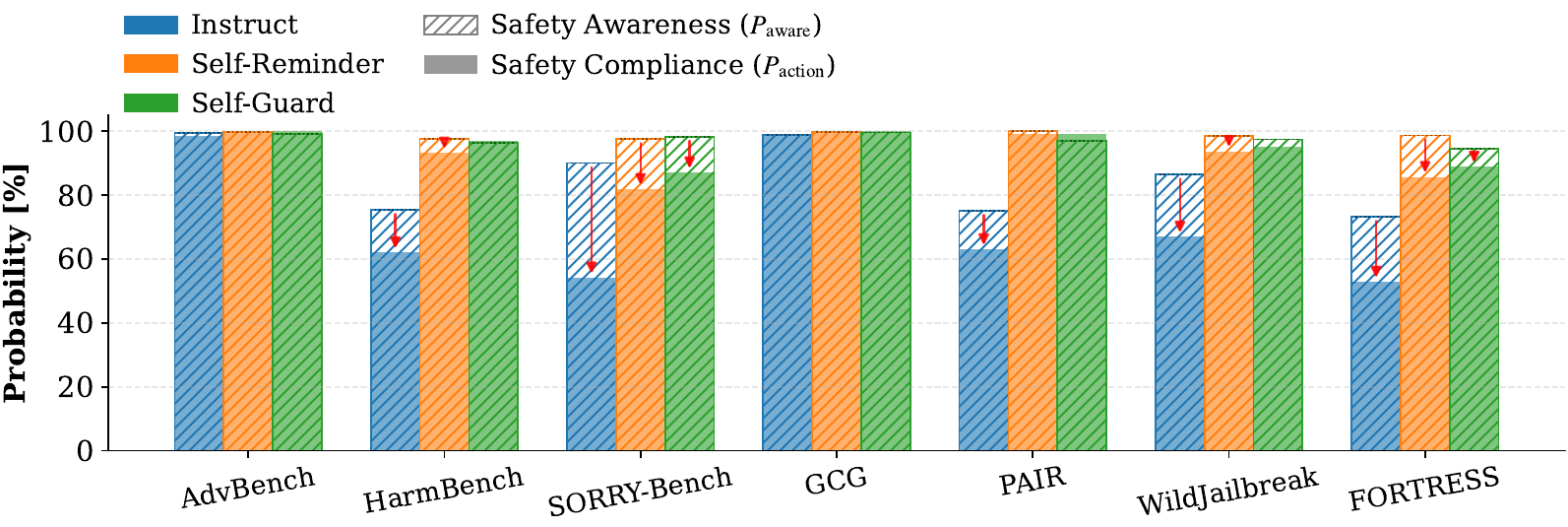}
    \caption{\textbf{\name narrows the Awareness-Compliance Gap on Qwen3-8B crosses full safety benchmarks.} Dashed bars represent Safety Awareness ($P_{\text{aware}}$), while solid bars represent Safety Compliance ($P_{\text{compliance}}$). Red arrows highlight the gaps where models detect risks but fail to refuse them.}
    \label{fig:app_8b_solo}
\end{figure}

\begin{figure}[h]
    \centering
    \includegraphics[width=\linewidth]{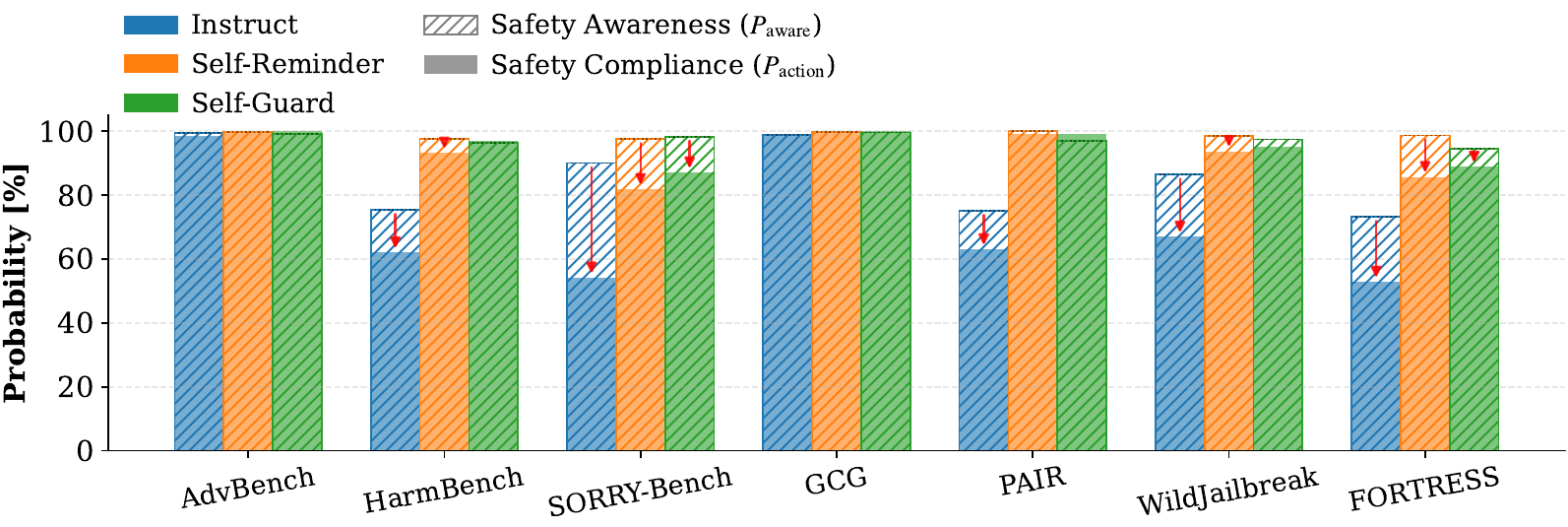}
    \caption{\textbf{\name narrows the Awareness-Compliance Gap on Qwen3-14B crosses full safety benchmarks.} Dashed bars represent Safety Awareness ($P_{\text{aware}}$), while solid bars represent Safety Compliance ($P_{\text{compliance}}$). Red arrows highlight the gaps where models detect risks but fail to refuse them.}
    \label{fig:app_14b_solo}
\end{figure}

We also report the details of the awareness-compliance gap for each model in Figure \ref{fig:app_4b_solo}-\ref{fig:app_14b_solo}. The results show that, compared with Self-Reminder (the safety prompt baseline), \name further steers models toward safety.

\section{Potential risk}

A potential risk of \name is that it may be abused for jailbreak attacks. Since \name operates by steering the activation space of LRMs toward safety behaviors, an adversary could intentionally reverse the sign of the steering strength to push the model away from safety. Such inverse steering may amplify unsafe or non-compliant behaviors, thereby facilitating jailbreak attempts.

\section{Offensive content declaration}

This paper contains examples of harmful, offensive, or toxic content (e.g., hate speech, jailbreak prompts) that serve as test cases for evaluating model safety. These examples are drawn directly from existing public datasets and benchmarks widely used in the safety research community (e.g., AdvBench, HarmBench, WildJailbreak, etc.). They are included solely to demonstrate the efficacy of the proposed methods and to advance the understanding of LLM vulnerabilities. We emphasize that this content does not reflect the authors' values or opinions.

\section{Data consent and authorization}

All data used in this work are obtained from publicly available datasets on HuggingFace\footnote{\url{https://huggingface.co}}. 
The dataset creators have provided official data consent documentation indicating that the data were collected and shared with appropriate consent. 
We completed their access questionnaires, received approval for research use, and used the datasets strictly in accordance with their licensing terms.

\section{Participants instructions}
\label{app:human_annotate}

To identify safety-related linguistic patterns, we employed a two-stage process involving LLM-based extraction followed by human verification. The full instructions provided to both the model and the human annotators are detailed below.

\subsection{Stage 1: LLM-based Token Extraction Instruction}
The following prompt was used to guide GPT in extracting candidate keywords from a large corpus of model responses:

\begin{table}[h]
\centering
\caption{Prompt for GPT5.2 for key word extraction.}
\label{tab:gpt_prompt}
\normalsize
\begin{tabular}{p{\linewidth}}
\toprule
Your task is to analyze the provided dataset of model responses and extract a set of representative tokens or phrases that indicate two specific behaviors. \\

1) Safety awareness: When the model acknowledges that a prompt may be illegal, harmful, or against policy. \\
2) Safety compliance: When the model explicitly declines to answer or provides a reasoned refusal. \\
\\
Return a list of keywords that serve as reliable indicators for these categories. \\
\bottomrule
\end{tabular}
\end{table}

\subsection{Stage 2: Human Verification Instruction}
After generating the candidate sets ($\mathcal{W}_{\text{pivot}}$ and $\mathcal{W}_{\text{refusal}}$), human annotators were given the following instructions to refine the sets:

\begin{table}[ht]
\centering
\caption{Instruction to human annotators}
\label{tab:human_prompt}
\normalsize
\begin{tabular}{p{\linewidth}}
\toprule

You are provided with two list of keywords extracted by ChatGPT. \\
For each set, you are given two tasks for evaluation. \\ 

\textbf{Task 1.} Precision Evaluation (token $\rightarrow$ indicator) \\
For a sample of 100 responses containing the keywords: check if the keyword actually indicates a safety-related behavior in its context. \\
Label as \textit{positive} if the keyword correctly identifies awareness/refusal, and \textit{negative} otherwise.

\textbf{Task 2.} Recall Evaluation (indicator $\rightarrow$ token)\\
For a sample of 100 raw model responses: 
Check if the current keyword set successfully captures the indication of awareness or refusal.\\
Label as \textit{positive} if the indication of awareness/refusal is captured by any keyword in the wordlist, and \textit{negative} otherwise.
\\
\bottomrule
\end{tabular}
\end{table}

To ensure objectivity, the verification task was performed independently by two annotators. 
The binary labels generated by both annotators for each evaluation task consistently exceeded a reliability threshold of 0.8 for Cohen's Kappa coefficient ($\kappa > 0.8$). 
This high level of inter-annotator agreement demonstrates that our collected keyword sets ($\mathcal{W}_{\text{pivot}}$ and $\mathcal{W}_{\text{refusal}}$) are statistically valid and robust indicators for identifying safety awareness and refusal behaviors.

\section{AI usage}

ChatGPT\footnote{https://chatgpt.com} and Gemini\footnote{https://gemini.google.com} were used to assist with debugging and language polishing.

\end{document}